\documentclass[conference]{IEEEtran}
\usepackage{times}
\usepackage{comment}

\usepackage[ruled,vlined]{algorithm2e}

\usepackage{bm}

\usepackage{graphicx}                       % Necessary to include graphics
\usepackage{graphics}                       % To include pdf, bitmapped graphics files
\usepackage{epsfig}                         % To include eps files
\usepackage[tight,footnotesize]{subfigure}  % Create subfigures, ie 1A, 1B
\graphicspath{./pics/RAL_fig_for_revise}
\graphicspath{./pics/RAL_new_fig}
\usepackage{amssymb,amsmath}%\usepackage{amslatex}%
\usepackage{mdwmath}
\usepackage{commath}   % Used for \norm \abs
\usepackage{eqparbox}
\usepackage{mathtools}
\usepackage[utf8]{inputenc} % Used for theorem/corollary/lemma
\usepackage[english]{babel}
 
\usepackage{stfloats}                       % Create Navigation Titles in PDF
\usepackage{url} % bibtex
\usepackage{hyperref}
\usepackage{cite}                           % Allows to establish ranges of papers for extended citations.
\usepackage[T1]{fontenc} % Underscores (to better visualize them using the appropriate font)
\usepackage{tabularx}
\usepackage{multirow}
\usepackage[table]{xcolor}
\usepackage{longtable}
\usepackage{booktabs} 			% professional-quality tables and \midrule
\usepackage{comment}
\usepackage{xcolor,colortbl}    % Color columns, use with: \usepackage{array}

\usepackage{nicematrix}
\usepackage{multicol}
\usepackage{graphicx}
\usepackage{placeins}
\usepackage{multirow}

\begin{document}

\title{One-Shot Domain-Adaptive Imitation Learning \\ via Progressive Learning}
\author{Dandan Zhang, Wen Fan, John Lloyd, Chenguang Yang, Nathan Lepora
\thanks{ Dandan Zhang Wen Fan, John Lloyd, Nathan Lepora are with the Department of Engineering Mathematics, University of Bristol;  Chenguang Yang is with the Department of Engineering Design and Mathematics, University of the West of England. All the authors are affiliated with Bristol Robotic Laboratory, UK.}}


\maketitle
\begin{abstract}
Traditional deep learning-based visual imitation learning techniques require a large amount of demonstration data for model training, and the pre-trained models are difficult to adapt to new scenarios. To address these limitations,
we propose a unified framework using a novel progressive learning approach comprised of three phases: i) a coarse learning phase for concept representation, ii) a fine learning phase for action generation, and iii) an imaginary learning phase for domain adaptation. Overall, this approach leads to a \textit{one-shot domain-adaptive imitation learning} framework. We use robotic pouring task as an example to evaluate its effectiveness. Our results show that the method has several advantages over contemporary end-to-end imitation learning approaches, including an improved success rate for task execution and more efficient training for deep imitation learning. In addition, the generalizability to new domains is improved, as demonstrated here with novel background, target container and granule combinations. We believe that the proposed method can be broadly applicable to different industrial or domestic applications that involve deep imitation learning for robotic manipulation, where the target scenarios have high diversity while the human demonstration data is limited.
\end{abstract}

%\IEEEpeerreviewmaketitle

\section{Introduction}

%,kulak2021active,zhang2021confidence,

Imitation learning is an effective tool for robots to learn dexterous manipulation skills \cite{wang2018masd,fu2019active,chen2020supervised,liu2022novel}, in scenarios where obtaining a dynamic model for control or specifying a reward function \cite{zhou2021manipulator}  for reinforcement learning are challenging. 
However, a large database is normally required for training control policies \cite{zhang2018deep}. Moreover, the policies trained in a specific environment may not work well if applied to other environments, due to the presence of domain gaps.  An ideal automatic robotic manipulation system should be able to adapt to new scenarios for task execution even if very limited demonstration data is available for training the control policy \cite{ma2022efficient}.  To this end, we develop a \textbf{one-shot domain adaptive imitation learning} framework that is data-efficient and can generalize the learned behavior to a new scenario with novel domain characteristics without significant loss in performance.

\begin{figure}[tb]
	\centering
	\includegraphics[width = 1\hsize]{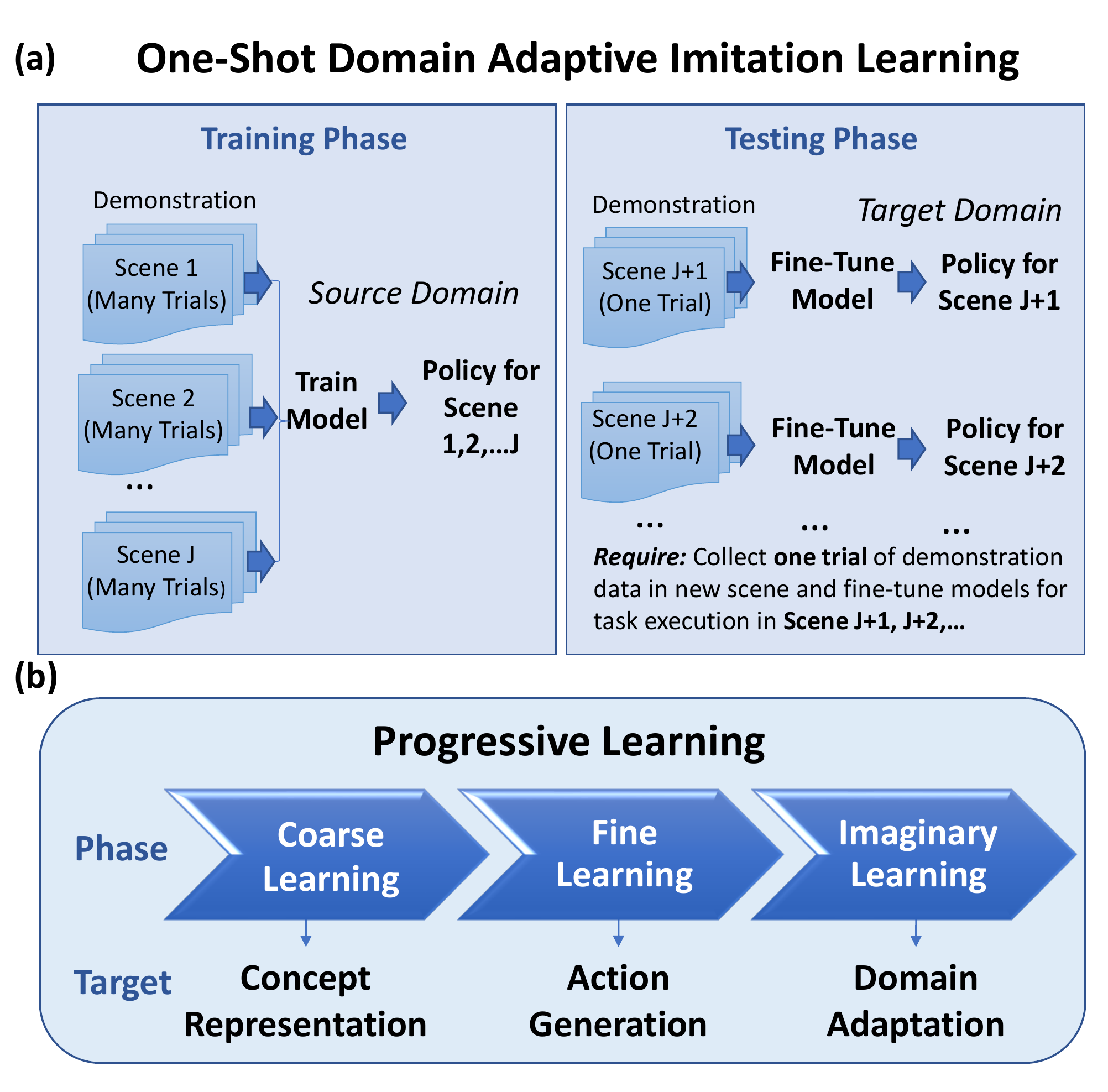}
	\caption{Schematic diagram of the proposed learning framework and approach. (a) Illustration of the one-shot domain adaptive imitation learning framework.  (b) The workflow of the progressive learning method, including a coarse learning phase, a fine learning phase, and an imaginary learning phase.  }
	\label{fig:SetupUR}
\end{figure}

%%%%%%%%%%%%%%%%%%%%%%%%%%%%%%%%%%%%%%
Humans are good at learning and generalizing strategies for everyday tasks from a few demonstrations. In particular, humans can learn the concepts for performing a series of tasks and then transfer that knowledge to new scenarios by practicing specific visuomotor skills.  Motivated by the advantages of human learners, we propose a \textbf{progressive learning} method that decouples the traditional end-to-end imitation learning pipeline into three phases: coarse learning, fine learning and imaginary learning.
Our central insight is to enforce the robot to acquire the general knowledge with a good concept representation in the coarse learning phase \cite{cheng2019purposive}, then learn to generate the precise motions in the fine learning phase, and finally expand this knowledge to new scenarios in the imaginary learning phase, which mimics the progressive learning process that humans also appear to do.   
 
We use robot pouring task as an example to evaluate the effectiveness of the proposed method, which is an essential skill for industrial or domestic robots when used for dispensing lubricants \cite{van2009medication}, carrying out chemical experiments \cite{saigal2008fast,kennedy2019autonomous}, cleaning \cite{pan2017feedback}, cooking \cite{neumann2017kognichef}.
We chose this task because pouring involves complex dynamic processes that are difficult to model~\cite{zhang2021explainable}. Moreover, it is not feasible for robots to learn from trial-and-error based on reinforcement learning approaches for the pouring task, because of the large amount of human intervention that would be required while training the robot. To this end, we consider the robotic pouring task as an appropriate example to validate the proposed imitation learning framework.

The  contributions of this paper are:\\
\noindent 1) We train the robot to learn general concepts by encoding concept representation features during the \textbf{coarse learning phase}, which provides compact but interpretable features extracted from raw pixels. This paves a way for the robot to learn action generation with high efficiency.\\
\noindent 2) We enable the robot to generate precise motion using an LSTM-Attention hybrid model during the \textbf{fine learning phase} based on the features extracted at the coarse learning phase, which ensures the success rate of task execution by incorporating concept representation with temporal information.\\
\noindent 3) We employ a generative adversarial network to generate a large amount of synthetic observation data in new scenarios during the \textbf{imaginary learning phase}, which enhances the perception skills of the robot and ensures that the robot can adapt the pre-trained policies to new scenarios with ease.

In summary, the \textbf{main contribution} of this paper is to formulate a one-shot domain adaptive imitation learning framework and demonstrate a progressive learning approach that can implement such framework. The proposed method can address the fundamental limitations of deep imitation learning by eliminating the need of recollecting a large amount of demonstration data and retraining the whole model in new domains with unseen object properties or environments. 

\section{Related Work}
\label{related}
\textbf{Imitation Learning.} 
Imitation learning considers the problem of acquiring skills from observing demonstrations.  Behaviour  Cloning and Inverse Reinforcement Learning (IRL)  are two major  research directions for imitation learning \cite{osa2018algorithmic,wang2018facilitating}. Behaviour Cloning aims to teach the robot to follow the expert guidance from supervised learning.  IRL estimates a reward function from human demonstration, and then the learned reward function is used for reinforcement learning. Survey articles  include \cite{argall2009survey,zheng2021imitation,fang2019survey}. 

 For traditional imitation learning, a large amount of demonstration data is normally required for training the policies \cite{zhang2018deep}.  Moreover, policies obtained during the model training phase are domain-specific \cite{nguyen2021most}. For example, a model trained for task execution in one scene through an imitation learning algorithm in the training phase may not have good performance for task execution in the testing phase when encountering a new scenario with new properties. Therefore, recent research for imitation learning has been focusing on one-shot learning and domain adaptation, which enhances the data efficiency and generalizability of traditional imitation learning approaches.

\textbf{ One-Shot Imitation Learning.} 
 One-shot imitation learning enables robots to learn to perform a new task with few demonstrations from humans  \cite{wang2020generalizing,vinyals2016matching,koch2015siamese,rezende2016one}. The objective of one-shot imitation learning is to train action prediction networks that are not specific to one task, then maximize the expected performance of the learned policy when faced with a new, previously-unseen task. In one example, a one-shot imitation learning method uses one demonstration and the observed state as the input and generates the action value for a 
block stacking task \cite{duan2017one}.  
The first vision-based one-shot imitation learning framework \cite{finn2017one} used Model-Agnostic Meta-Learning (MAML) \cite{finn2017model} for an object placing task.
 However, this method requires the training database to have high diversity with many demonstration trials for different tasks; for example, about 1300 demonstrations were collected for meta-training during the training phase \cite{finn2017one}.  Therefore, we aim to eliminate the need of collecting a large amount of diverse demonstration data during the training phase in this paper. Moreover, to verify that the one-shot learning can be implemented for more complex tasks, we evaluate our proposed method on a robotic pouring task, which involves more complex dynamic processes.
 
 % high-fidelity trajectory imitation

\textbf{Domain Adaptation.}
Domain adaptation is a process that allows a neural network model trained with samples from a source domain to generalize to a target domain \cite{bousmalis2018using}.
Recent domain adaptation methods learn deep neural transformations that map image data from dstinct domains into a common feature space. For example, adversarial discriminative domain adaption (ADDA) \cite{tzeng2017adversarial} is a method for domain adaptation in image classification. However, the labels in the source domain and the target domain were required to be identical, which is unrealistic for imitation learning in robotics and does not align with the one-shot imitation learning setting. 
Another research direction is to 
reconstruct the target domain from the source representation \cite{ghifary2016deep}. To this end, a generative adversarial network can be used to generate a large database with synthetic observation data for model training to support domain adaptation \cite{bousmalis2017unsupervised}, which we will use in this paper.

\textbf{Learning-Based Robotic Pouring.} 
Previously, one imitation learning approach has been investigated for a dynamic fluid pouring task similar to the one in this paper \cite{langsfeld2014incorporating}. However, that previous study required many failed human demonstrations for the robot to learn how to recover from errors \cite{langsfeld2014incorporating}. In another work, an RNN-enabled MPC (Model Predictive Control) controller determined the optimal velocity for execution, which was verified with a custom apparatus. A single motor was used to generate the rotational motions \cite{chen2019accurate}, instead of a robotic arm. While the approach was able to learning the task, the generalizability was not  demonstrated, in particular the 3D position of the source container cannot be adjusted.
Thus, the above methods can perform well in specific scenarios, but there are open questions about how well they would carry over to new scenarios with novel target containers, granules and backgrounds. To this end,  learning-based method for robotic pouring task with generalizability will be developed and evaluated in this paper.

\section{Background}
\label{Methodology}

\subsection{Overview}
 Our goal is to learn a policy that can  adapt to new domains from a single demonstration of that task during the testing phase, while eliminating the need of collecting a large amount of data from different scenes during the training phase (contrasting with traditional one-shot learning, as discussed above). As shown in Fig. \ref{fig:SetupUR}(a), the data collection process is similar to that of traditional imitation learning, with several trials of demonstration data collected for several scenes (Scene 1, 2,$\cdots$, $J$). During the testing phase, for task execution in Scene $J+1$, $J+2$, $\cdots$, only one trial of demonstration data is required. The model obtained in the training phase can be fine-tuned during the testing phase and new policies for task execution in new scenes can be obtained quickly. 
 
 Our approach is to develop a progressive learning approach, which may be considered as a representative architecture for the one-shot domain-adaptive imitation learning scheme. The workflow of our progressive learning approach is shown in Fig. \ref{fig:SetupUR}(b).

\subsection{Problem Formulation}

We use $\boldsymbol{o_t}$ to represent  the observation and $\boldsymbol{a_t}$  to represent the action  at time $t$.   Then ${\tau}$ represents a trajectory for performing the task, consisting a sequence of observation and action pairs:
${\tau} = \left[\left(\boldsymbol{o}_{1}, \boldsymbol{a}_{1}\right), (\boldsymbol{o}_{2},\boldsymbol{a}_{2}), \ldots,\left(\boldsymbol{o}_{T}, \boldsymbol{a}_{T}\right)\right]$.

\textbf{Training Phase.}
Let $\mathcal{T}_j=\left\{\tau_{1},\tau_{2}, \cdots,  \tau_{K}\right\}$ denotes a group of trajectories for task execution in Scene $j$.
Let $\mathcal{D}_{\mathcal{T}} = \{T_1, T_2, \cdots, T_J\}$ denotes the database for model training, which is comprised of different trajectories $\tau_{k}(k=1,2, \cdots, K)$ for task execution in different scenes $T_j(j=1,2, \cdots, J)$, demonstrated by human for the robot to imitate.  A policy $\boldsymbol{\pi}_{{\Theta}}$ will be obtained to map observations $\boldsymbol{o_t}(t=1,2,...,T)$ to actions $\boldsymbol{a_t}(t=1,2,...,T)$ under the parameters ${\Theta}$. We denote the observation data used for model training in the training phase as being from source domain $\bm{H}$.

\textbf{Testing Phase.}
Let $\mathcal{D}_{\mathcal{T}}^{r} = \{T_{J+1}, T_{J+2}, \cdots\}$ denote the new database collected during the testing phase. $T_{J+1}, T_{J+2}, \cdots$ represent new scenarios for task execution, which have novel domain characteristics unseen during the training phase. Unlike the training phase (where each scene $T_{j}$ has several demonstrated trajectories), $\mathcal{D}_{\mathcal{T}}^{r}$ only includes a single  trajectory collected as demonstration data for each scene $T_{J+1}, T_{J+2}, \cdots$, while the corresponding policies ${\pi}_{{\Theta}_{J+1}}$, ${\pi}_{{\Theta}_{J+2}}$, $\cdots$ for task execution are obtained via transfer learning.
We denote the observation data collected during the testing phase as being from target domain $\bm{R}$.

\subsection{Robotic Pouring Task}
Here we use robotic pouring as a concrete example to demonstrate the proposed progressive learning approach. 

\textbf{Task Description.}
The pouring task requires the robot to successfully pour different granular materials to different target containers with distinct background environments. More specifically, the robot should learn to adjust a reasonable position of the source container and control the wrist motions in a proper manner to avoid spilling the materials out of the  target containers.  Here we consider distinct scenes to correspond to cases where either granular materials, target containers or backgrounds are different from previously experienced combinations. We collected human demonstration data to train a model for online deployment. The learnt model will generate action values to command the robot to execute the pouring task. The robot then returns to its original starting configuration after each episode of the pouring task is completed.

%For simplicity, we will assume that the robot approaches the target container before task execution using a standard control method, then t

\begin{figure}[tb]
	\centering
	\includegraphics[width = 1\hsize]{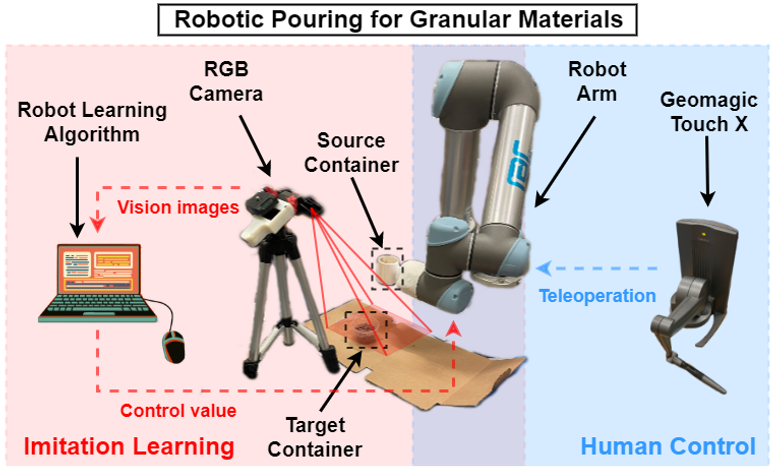}
	\caption{The  experimental setup for the robotic  pouring task. A pouring container is mounted as the end effector to the wrist of a UR5 (Universal Robots) 6-axis robot arm. An RGB camera is mounted at a fixed position to view a target container placed on a table adjacent to the base of the arm. Human demonstration data is collected via teleoperation using Geomagic Touch X haptic motion-capture device as the remote controller. }
	\label{fig:SetupUR1}
\end{figure}

\textbf{Hardware Deployment.} The human demonstration database for model training was collected via teleoperation of a UR5 (Universal Robots) 6-axis robot arm, using a Geomteric Touch X (3D Systems) haptic motion-capture device as the remote controller to provide human-guided commands. An RGB camera sited to view the container was used to capture the image frames for training. The source container was attached as an end effect to the wrist of the robotic arm, with the target container placed on a table near the base of the robot arm. The  experimental setup for the robotic  pouring task is shown in Fig. \ref{fig:SetupUR1}.

\subsection{Database Construction}

\begin{table*}[tb]
	\caption{\textsc{Details of the Database}}
	\label{tab:Data1}
	\centering
	\begin{tabular}{c|c|c|c|c|c|c|c|c|c|c|c|l}
		\hline
		\multirow{2}{*}{\textbf{Scene}} &	\multicolumn{4}{c|}{\textbf{Target Container}} &  \multicolumn{3}{c|}{\textbf{Granules}}& \multicolumn{2}{c|}{\textbf{Background}} & \multicolumn{2}{c|}{\textbf{Trails}}&\multirow{2}{*}{~~~\textbf{Evaluation}} \\\cline{2-12}
		& \textbf{Type}  & \textbf{Property}  & \textbf{Size} & \textbf{Color} & \textbf{Type} & \textbf{Color}  & \textbf{Shape}&\textbf{Color} & \textbf{Type} & \textbf{Train}  & \textbf{Test}   \\\hline
		\textbf{1}  & Goblet & Opaque &  Medium &White  & Lentils & Green  & Oblate& Orange&
		Board&8 & 1 \\
		\textbf{2}  & Goblet  & Opaque &   Medium&White  & Rice & White & Prolate & Orange &
		Board&8 & 1 \\
		\textbf{3}  & Goblet  & Opaque &   Medium&White &  Couscous & Yellow & Cylindrical  &Orange&
		Board& 8 & 1 & $\checkmark$\\
		\textbf{4}  &  Plate  & Opaque & Small&White  & Lentils & Green & Oblate &Orange &
		Board&8 & 1 & $\checkmark$\\
		\textbf{5}  & Plate  & Opaque & Small&White &  Rice & White & Prolate &Orange&
		Board& 8 & 1 \\			
		\textbf{6}  & Plate  & Opaque & Small&White &  Couscous & Yellow   & Cylindrical &Orange&
		Board& 8 & 1 \\		
		\textbf{7}  &Cup  & Opaque & Small&White  &  Lentils & Green & Oblate &Orange&
		Board& 8 & 1 \\
		\textbf{8}  & Cup  & Opaque  & Small&White &    Rice & White & Prolate &Orange&
		Board& 8 & 1 & $\checkmark$\\
		\textbf{9}  &  Jar   & Transparent & Big&/ &   Lentils & Green & Oblate &Orange&
		Board& 8 & 1 \\
		\textbf{10}  & Jar & Transparent &  Big&/ & Rice & White & Prolate &Orange&
		Board& 8 & 1 & $\checkmark$ \\\hline	 
		
		\textbf{11}  & Cup & Opaque &  Small&White & Couscous & Yellow   & Cylindrical &Orange&
		Board& 0 & 1 &   \\
		\textbf{12}  &  Jar   & Transparent & Big&/ &   Couscous & Yellow   & Cylindrical &Orange&
		Board& 0 & 1 &   \\
		\textbf{13}  &  Jar   & Transparent & Big&/ &   Lentils & Green   & Oblate &Blue&
		Tissue& 0 & 1 & $\checkmark$(New Background) \\
		\textbf{14}  & Plate  & Opaque & Small&White & Lentils & Green   & Oblate & Blue &
		Tissue & 0 & 1 & ~~(New Background)\\	 	
		\textbf{15}  & Goblet  & Opaque &   Medium&White  & Lentils & Green   &  Oblate & Blue &
		Tissue & 0 & 1 & ~~(New Background)\\	
		\textbf{16}  &  Cup   & Opaque & Small&White &  Coffee & Brown   & Irregular &Orange&
		Board& 0 & 1 & $\checkmark$(New Granules) \\
		\textbf{17}  & Plate  & Opaque & Small&White & Coffee & Brown   & Irregular & Orange&
		Board& 0 & 1 & ~~(New Granules)\\
		\textbf{18}  & Goblet  & Opaque &   Medium&White  & Coffee & Brown   & Irregular &Orange&
		Board & 0 & 1 & ~~(New Granules)\\			\textbf{19}  & Cup  & Opaque & Big&Black & Rice & White & Prolate & Orange&
		Board & 0 & 1 & $\checkmark$(New Container)\\
		\textbf{20}  & Cup  & Opaque & Big&Black  & Couscous & Yellow   & Cylindrical &Orange&
		Board & 0 & 1 & ~~(New Container)\\ 
		\hline
	\end{tabular}
	$\checkmark$ represents that the scene will be used for physical experiments for model evaluation.
\end{table*}

\begin{figure}[tb]
	\centering
	\includegraphics[width = 1\hsize]{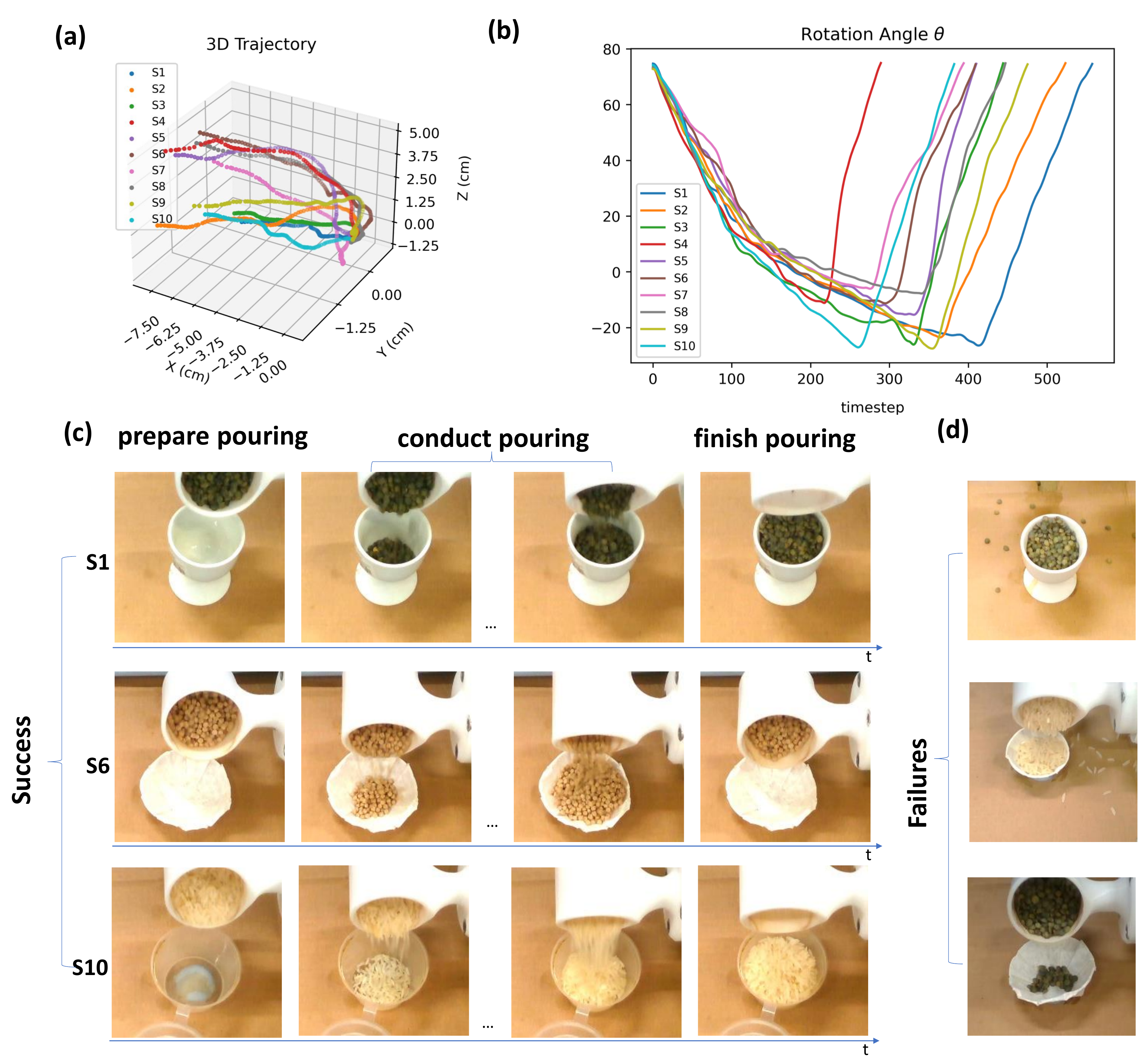}
	\caption{Overview of the constructed database for model training. (a)  Raw trajectories of the 3D position of the end-effector. (`S' is the abbreviation of `Scene') (b) Raw trajectories of the rotation angle of the robot's wrist joint. (c) Examples of three successful trails of the pouring experiments.  (d) Examples of failed trails. }
	\label{tab:Data}
\end{figure}

\textbf{Performance Evaluation.}
The success rate is used to evaluate the performance of the imitation learning algorithm, which is a ratio between the number of successful trials and the total conducted during the experiments. A successful trial is defined as one in which the granules are poured from the source container into the target container without spilling. If the total volume of the granules in the source container is smaller than the maximum capacity of the target container, a successful trial is defined as pouring at least 90\% total volume of granules from the source container to the target container. If the maximum capacity of the target container is smaller than the source container, the target container should be filled to at least 90\% of the total capacity of the target container. 

Ten raw trajectories, including the 3D position and the rotation angle of the wrist joint of the robot's end-effector, are plotted in Fig. \ref{tab:Data}(a)-(b) as examples. 
Fig. \ref{tab:Data}(c) shows three successful trials of robotic pouring as examples, while Fig. \ref{tab:Data}(d) indicates failures when executing the pouring task.

\begin{figure}[tb]
	\centering
	\includegraphics[width = 1\hsize]{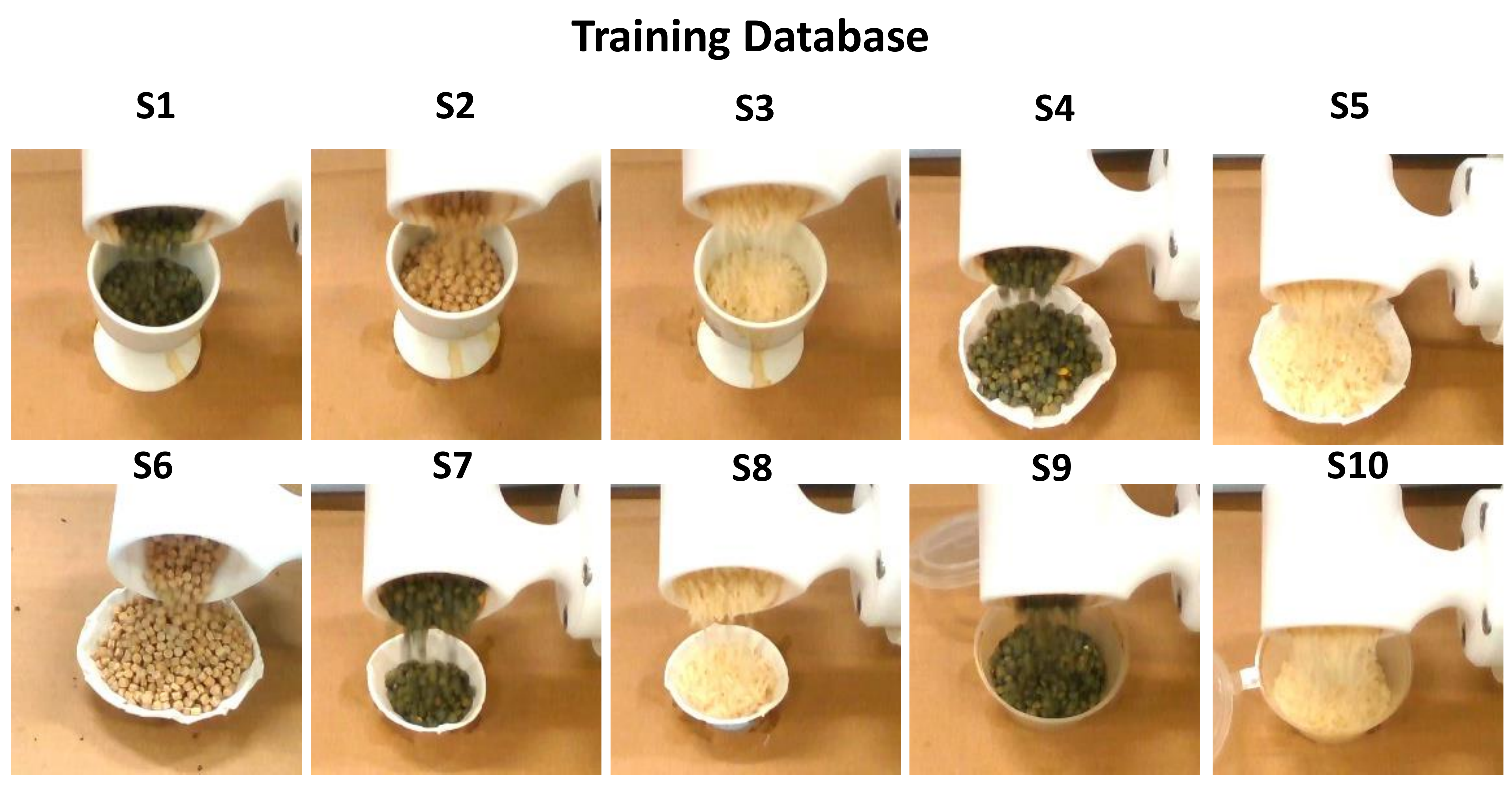}
	\caption{Examples of the ten different pouring scenes, S1-S10, for constructing the training database (`S' abbreviates `Scene').}
	\label{tab:Dat1a11}
\end{figure}

The demonstration database is constructed from ten distinct pouring scenes (known as scene S1-S10).  Eight trials are collected for each scene respectively, as shown in Fig. \ref{tab:Dat1a11}. The properties of the target containers, granules and background used for different pouring scene are summarized in Table \ref{tab:Data1}.
Four different types of target containers are used for experiments: a `goblet', `plate', `cup' and `jar' that have various properties, sizes, and colors.  As for the granules, we used `lentils', `rice', and `couscous', which have different colors and shapes. A sheet of orange cardboard is used as the background during data collection, which will be changed during testing.

The database contains a series of trajectories that are comprised of observation-action pairs.  This training database, denoted as $\mathcal{D}_{\mathcal{T}}$, will be used for model training in the coarse learning phase and fine learning phase. The images captured as observation  data $\bm{o_t}$  during teleoperation are cropped to size of $224\times224$ at 30 fps. The robot kinematics states are recorded simultaneously as the corresponding action values. To pour the granules, the tilt angle of the source container is controlled directly by commanding the rotating angle $\theta(t)$ of the wrist joint of the robotic arm, while the 3-dimensional (3D) position $p(t) = [x(t),y(t),z(t)]$ of the end-effector is adjusted to ensure pouring without spillage. We use the 3D velocity $v(t) = [v_x(t),v_y(t),v_z(t)]$ and rotation angle  for robot control during online deployment. Therefore, we define action values as  $\boldsymbol{a_t}=[v_x(t),v_y(t),v_z(t),\theta(t)]$ in this paper.

\section{Methodology: Progressive Learning}
\begin{figure*}[tb]
	\centering
	\includegraphics[width = 1\hsize]{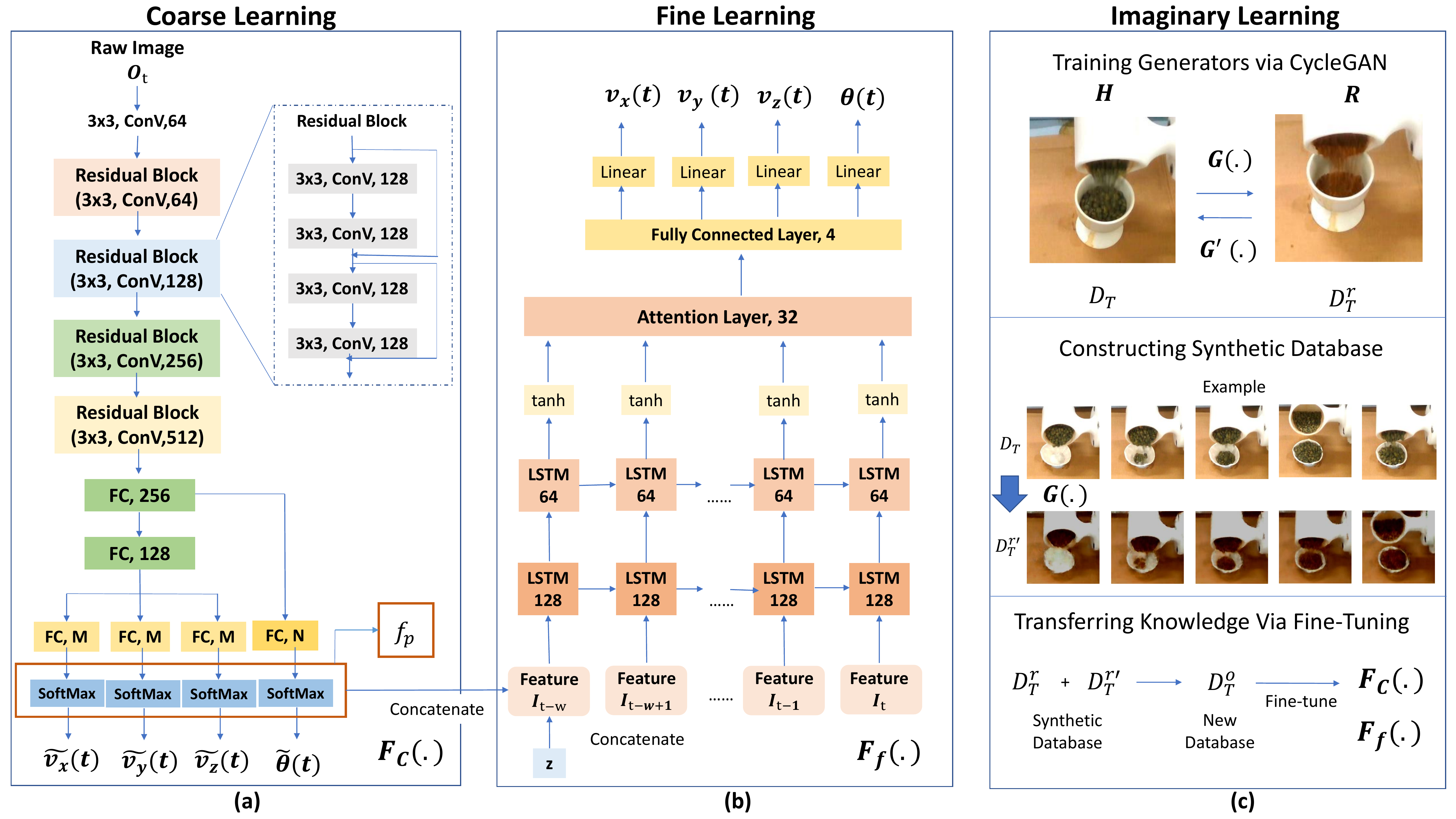}
	\caption{The architecture of the progressive learning method.  (a) The architecture of the  coarse learning model for feature extraction, (b) The architecture of the  fine learning model for action generation, (c) The workflow illustration of the imaginary phase for domain adaptation.}
	\label{fig:Framework}
\end{figure*}

The neural network architecture of the progressive imitation learning method is shown in Fig. \ref{fig:Framework}.
\subsection{\textbf{Coarse Learning}: Concept Representation}

The main goal of the coarse learning phase is to enable the robot to learn basic concepts by encoding representation from raw pixels, which can be used to accelerate the action generation model training during the fine learning phase while ensuring model performance by preserving compact but interpretable features.

 \subsubsection{Design Consideration}

Representation learning techniques aim to extract features from high-dimensional sensory input, which can help improve the model performance in some downstream learning tasks \cite{pari2021surprising}.
Though deep neural networks such as 
 VGG16, InceptionV3, and ResNet50 \cite{theckedath2020detecting} with pre-trained weights on ImageNet can be used for transfer learning, the encoded features cannot explicitly express the contexts for a specific task. Auto-encoders and their variances can be used for feature extraction, but these models cannot guarantee the representation is human-interpretable.  Therefore, here we aim to train a model that can learn concept  representation  with interpretable features that benefit the downstream process (action generation) when data is limited.

\subsubsection{Implementation}

The concept representation of the pouring task is comprised of three key components: i) the tilt angle control, ii) the 3D-position adjustment, and iii) the task-specific characteristics (encoded as a distinct variable $z$).

\textbf{\textit{Component 1:}} \textbf{Tilt Angle Control.} Imagine that a `teacher' (human demonstrator) is responsible for supervising a `student' (robot imitator) to learn the pouring task by transferring general concepts using descriptive language. 
The whole pouring process could then be split into several stages, such as:\\
\noindent (1) increase the tilt angle of the source container quickly until granules flow out from the source container;\\
\noindent (2) keep increasing the  tilt angle of the source container stably to fill up the target container;\\
\noindent (3) reduce the tilt angle of the source container when the target container is almost full or no remaining granules are in the source container.

 Suppose that ${\theta}_s$ is the tilt angle of  the source container when the granules begin to leave the container. Then ${\theta}_m$ represents the maximum tilt angle, after which the robot starts to restore to its original pose.  The tilt angle of the source container is the rotation angle of the robot's wrist (axis-6), denoted by $\theta(t)$.

For a simple implementation, the first stage of the task could be labelled as $\widetilde{\theta}(t)=0$ (when ${\theta}(t) < {\theta}_s$), and the second and the third stages labelled as $\widetilde{\theta}(t)=1$ (when ${\theta}_s  \leq {\theta}(t) \leq {\theta}_m$) and $\widetilde{\theta}(t) = 2$ (when ${\theta}_m < {\theta}(t)$) respectively. The supervision for tilt angle control could then be formulated as a 3-class classification problem on the visual images.

For a more general implementation, we want the `student' to learn  detailed knowledge, such as `increase  the tilt angle slightly' or `increase the tilt angle significantly'. In this case, the second stage could be further segmented into many sub-stages. Suppose that there are $N$ stages in total with the first stage is denoted as $\widetilde{\theta}(t) = 0$ and the final stage denoted as $\widetilde{\theta}(t) = N-1$. Then, we define the discrete class labels as
\begin{equation}
	\label{theta}
\widetilde{\theta}(t)  =\left\{
	\begin{array}{ll}
	0 &  \mathrm{if}~{\theta}(t) < {\theta}_s
	     \\
		\lceil ( {\theta}(t) - {\theta}_s)  / {\theta}_r \rceil   & \mathrm{if}~  {\theta}_s  \leq {\theta}(t) \leq {\theta}_m \\
		N-1 & \mathrm{if}~   {\theta}_m < {\theta}(t),
	\end{array} \right. 
\end{equation}
where ${\theta}_r= ({\theta}_s-{\theta}_m)/(N-2)$ represents a normalization to the original simpler implementation during the second stage. In consequence, the class label $\widetilde{\theta}(t)$ for a specific observation $o_t$ is integer-valued, and the supervision for tilt angle control is formulated as an N-class classification problem.

\textbf{\textit{Component 2:} } \textbf{3D Position Adjustment.}
During each stage, to avoid the granules missing the target container, the `teacher' should instruct the `student' to adjust the 3D position of the source container in a reasonable manner. For example, if the source container is far away from the target container, the `teacher' can give instructions to the student like `move forwards', `move backwards', `move left', `move right', `move up', `move down' to control the source container to reach a desired position for pouring that depends on the relative 3D-position between the two containers. If the source container has already been located at a reasonable position, the instruction can be simply to `keep still'.

Suppose that $v_q(t)(q=x,y,z)$ represent the end-effector's linear velocity along the $x$-, $y$- and $z$-axis respectively.  Then we denote ${v}_s>0$ as a threshold value on intentional motion, such that if $||v_q(t)||<{v}_s$, then the velocity is regarded as not caused by a human's intention to move but by unintentional motion, e.g. tremor, during the teleoperation.
For the simplest implementation,  $\widetilde{v}_{q}(t)=0$ when ${v}_s\leq v_q(t)$ and $\widetilde{v}_{q}(t)=2$ when $v_q(t)\leq-{v}_s$, representing motions in the positive and negative directions along the $x$-, $y$-, or $z$-axis respectively; meanwhile, $\widetilde{v}_{q}(t)=1$ when $||v_q(t)||<{v}_s$ represents not moving. In this case, the supervision for 3D-position control can be formulated as a 3-class classification problem.

%the mapping from $v(q)$ to $	\widetilde{v}_{q}(t)$ can be obtained as follows.

For a more general implementation, we want the `student' to learn both the  direction and  magnitude for 3D-position adjustment, such as to `move quickly in the positive direction' or `move slowly in the positive direction'. Suppose that there are $M$-classes of instructions for position adjustment in total, then the
class labels are defined as:
\begin{equation}
	\label{xyz}
	\widetilde{v}_{q}(t) =\left\{
	\begin{array}{ll}
		0 & \mathrm{if}~ {v}_s \leq v_{q}(t)\\
		M-1 & \mathrm{if}~  v_{q}(t)\leq -{v}_s \\
		\lceil(v_{q}(t) + {v}_s)/m \rceil & \mathrm{if}~-{v}_s<v_{q}(t)<{v}_s 
	\end{array} \right. 
\end{equation}
where $m = 2{v}_s/(M-2)$ is a normalization to give integer class labels.

\textbf{\textit{Component 3:}} \textbf{Task-Specific Characteristics Encoding.} In addition, a `teacher' may also give task-specific information to the `student' that could include physical factors that affect the pouring dynamics. For example, if the granules/liquids used for pouring have high friction, then the pouring velocity should be higher to ensure they flow quickly out of the source container. Conversely, if the target container is small in size, then the velocity for adjusting the 3D-position of the source container should be relatively slow to minimize spilling the granules/liquids out of the target container.

Here a variable $z$ is used to represent the characteristics of the task, relating to factors that affect the pouring dynamics but do not change with time. In this work, $z$ = $[C,P]$, where $C$ is a scalar that indicates the size of the target container (depending on its maximum capacity) and $P$ is a scalar that indicates the type of granules used for pouring. Here we used three distinct types of granules in the experiments, denoted by 3 scalar values ($P$=0,1,2 for lentils, rice and couscous).

%For example, we have four different types of target containers used in  experiments. We  therefore allocate 4 scalar ($C$=0,1,2,3) for different target containers. Similarly

\textbf{Concept Representation.} The goal of the coarse learning phase is to obtain a concept representation feature vector that includes information from the three components (see above) in an interpretable manner. Since the first and second component can be formulated as a classification problem, we use categorical cross-entropy loss ${L}_p(p=0,1,2,3)$ to update the parameters of the neural network model by maximizing the accuracy of prediction of the distinct variables $\widetilde{\theta}(t)$, $\widetilde{v_x}(t)$, $\widetilde{v_y}(t)$, $\widetilde{v_z}(t)$. The overall loss function $L$ is the linear combination of ${L}_p$ with different weights. For the pouring task, tilt angle control is more important and has higher relationship with the success rate of task execution. We set the weight as 0.4,0.2,0.2,0.2 for ${L}_p$ respectively in this paper.

During model training, we draw a batch of samples $\{X_k,Y_k\}(k=1,2,...,K)$ to update the parameters of $\bm{F_C(.)}$ in a supervised learning manner, with batch spize $K$, inputs $X_k = \bm{o_t}$ and outputs $Y_k=[\widetilde{v_x}(t),\widetilde{v_y}(t),\widetilde{v_z}(t), \widetilde{\theta}(t)]$.

When the model training is done, a “softmax” output layer is then used to learn each of the distinct outputs. After obtaining the probability distribution of each output variable,  all four probability distributions are concatenated into a single feature vector $f_p(t)$ of dimension $3M + N$. Subsequently, this feature vector $f_p(t)$ is concatenated with the contextual representation $z$ to give an overall feature vector $V_F^{t}=[f_p(t),z]$, which is obtained for each image frame $\bm{o_t}$.

Here we will use the class probabilities produced by the coarse  learning model as `soft inputs' to train the fine learning model (see below), instead of using the `hard inputs' (one-hot values) \cite{hinton2015distilling}. Our reasoning is that the `soft inputs' have high entropy and thus contain more information than `hard inputs', because they not only provide information on the most probable class but also on the other classes according to their probabilities.
 
 The neural network model for coarse learning can be trained via self-supervision, since  the labels  can be generated automatically based on \eqref{theta},\eqref{xyz}. We don’t need to label the data for concept representation in the coarse learning phase.

\subsubsection{Architecture}
The architecture of the coarse learning model, denoted as $\bm{F_C(.)}$, is an adapted version of a ResNet-18 model \cite{He_2016_CVPR}. Different from the original  ResNet-18 model for single image classification, we reorganize the architecture into a multi-head structure for multi-variable classification.

After $\bm{F_C(.)}$ is trained using database $\mathcal{D}_{\mathcal{T}}$, we use it to convert the image data to their corresponding concept representation feature vectors.  A new database $\mathcal{D}'$  is then constructed  in which the original observation-action pairs
${\tau}_k = \left[\left(\boldsymbol{o}_{1}, \boldsymbol{a}_{1}\right), (\boldsymbol{o}_{2},\boldsymbol{a}_{2}), \ldots,\left(\boldsymbol{o}_{T}, \boldsymbol{a}_{T}\right)\right]$ are replaced by state-action pairs ${\tau}_k' = \left[\left(\boldsymbol{s}_{1}, \boldsymbol{a}_{1}\right), (\boldsymbol{s}_{2},\boldsymbol{a}_{2}), \ldots,\left(\boldsymbol{s}_{T}, \boldsymbol{a}_{T}\right)\right]$ with $\bm{s_t} = V_F^{t}$. This database will be used for the fine learning process next.

\begin{algorithm}
	\SetAlgoLined	
	\textbf{Input:} Demonstration Database $\mathcal{D}_{\mathcal{T}}$\\
	\textbf{Required:}  	
	learning rate $\alpha$; batch size $K$; \\

\textbf{Coarse Learning Phase};\\
Initialize parameters ${\Theta}$ for $\bm{F_{C}(.)}$; \\
	\While{Training $\bm{F_{C}}({\Theta},\mathcal{D}_{\mathcal{T}})$}{
Sample batch of trajectories from  $\mathcal{D}_{\mathcal{T}}$;\\
Convert  $\theta(t)$ to $\widetilde{\theta}(t)$ based on  \eqref{theta}; \\
Convert $v_{q}(t)$ to $\widetilde{v}_{q}(t)$ based on \eqref{xyz};\\
Construct $\{X_k,Y_k\}(k=1,2,...,K)$ for training;\\
Compute loss function $L$;\\
Backpropagate gradient of $\bm{F_{C}(.)}$;\\
	Update parameters ${\Theta}$;}

	Construct new database $\mathcal{D}^{'}$ using $\bm{F_{C}(.)}$;\\

\textbf{Fine Learning Phase};\\
	Initialize parameters ${\Theta}_r$ of $\bm{F_{f}(.)}$;\\
	\While{Training $\bm{F_{f}}({\Theta}_r,\mathcal{D}^{'})$}
	{
		Sample batch of trajectories ${\tau}_k$ from  $\mathcal{D}^{'}$;\\
	Construct sequential data based on ${\tau}_k$:	$\bm{V_F^{w}} = [V_F^{t-w+1}, V_F^{t-w+2}, ..., V_F^{t-1}, V_F^{t}]$;\\
Construct $\{X_k,Y_k\}(k=1,2,...,K)$ for training;\\
Compute loss function $\mathcal{L}_{r}$;\\
Backpropagate gradient of $\bm{F_{f}(.)}$; \\
Update parameter ${\Theta}_r$;
	}

\textbf{Imaginary Learning Phase};

Construct database $\mathcal{D}_{\mathcal{T}}^{r}$ with one-shot demonstration; \\
Draw observation data from $\mathcal{D}_{\mathcal{T}}$  as source domain $\bm{H}$; \\
Draw observation data from $\mathcal{D}_{\mathcal{T}}^{r}$ as target domain $\bm{R}$; \\ 
Obtain the optimal $\bm{G(.)}$ based on \eqref{min-max}; \\(see \textbf{Appendix})\\
Sample ${\tau}_{j(j=1,2,\cdots)}$=$[(\bm{o_1},\bm{a_1}),(\bm{o_2},\bm{a_2}), {\cdots}]$;\\
Generate new trajectories: ${\tau}^{'}_j(j=1,2,\cdots)=[(\bm{G}(\bm{o_1}),\bm{a_1}),(\bm{G}(\bm{o_2}),\bm{a_2}), \cdots ]$;\\
Construct $\mathcal{D}_{\mathcal{T}}^{r'}$ with ${\tau}^{'}_j(j=1,2,\cdots)$;\\
$\mathcal{D}_{\mathcal{T}}^{o}$ $\leftarrow$ $\mathcal{D}_{\mathcal{T}}^{r}$ $\cup$ $\mathcal{D}_{\mathcal{T}}^{r'}$;

	Fine-tune  $\bm{F_C(.)}$ using $\mathcal{D}_{\mathcal{T}}^{o}$;\\
	Construct new database $\mathcal{D}_{\mathcal{T}}^{o'}$;\\
	Fine-tune the parameters of $\bm{F_{f}(.)}$ using $\mathcal{D}_{\mathcal{T}}^{o'}$;\\

	\textbf{Output:} Model $\bm{F_{C}(.)}$,$\bm{F_{f}(.)}$. %$[{\omega}_{min}, {\omega}_{max}] (s=1,2,\dotsc,N)$\\
	\caption{Progressive Learning}
\end{algorithm}

\subsection{\textbf{Fine Learning}: Action Generation}
 \subsubsection{Design Consideration}
The fine learning process aims to utilize the general concepts obtained by the coarse learning phase to generate the rotation angle of the robotic end-effector (pouring container) and the 3D linear velocity for position adjustments during the pouring process. 

We use a Long-Short Term Memory (LSTM) recurrent neural network \cite{Khatib1998Inertial}, which  contains memory cells and gates that  allow the network to propagate gradients back in time.  The network takes the inputs at the current time step alongside hidden states from previous time steps to generate the output for the current time step. LSTM  models are good at  processing sequential data,  and will thus be used as an essential architecture to construct a fine learning model of the action values.

A potential issue with the LSTM model is that it may overfit by memorizing the mean trajectory, which makes it harder to generalize to novel tasks and thus limits its generalizability. To address this issue with the vanilla LSTM architecture, we incorporate an attention mechanism \cite{vaswani2017attention}: instead of considering all neighbors as equal, the neighboring neurons are weighted according to a criterion specified by the attention model.

\subsubsection{Implementation}

After obtaining the extracted concept representation feature vector of each frame, a sequence of feature vectors can be formed by $\bm{V_F^{w}} = [V_F^{t-w+1}, V_F^{t-w+2}, ..., V_F^{t-1}, V_F^{t}]$,  where the parameter $w$ sets the length of the sequence.   The fine learning model uses the features obtained from observation images at these times $t-w+1$,  $t-w+2$, $\dots$, $t-1$, $t$ as input, with the rotation angle and 3D linear velocity of the robot’s end-effector as output for robot control. During model training, we draw a batch of samples $\{X_k,Y_k\}(k=1,2,...,K)$ to update the parameters of $\bm{F_f(.)}$ in a supervised learning manner, where $K$ represents the batch size and$X_k = \bm{V_F^{w}}$, $Y_k=[{\theta}(t),{v_x}(t),{v_y}(t),{v_z}(t)]$.

To ensure that the generated velocities are safe and reasonable for the controlling the robot, we calculate the mean and variance of the angular velocity ${\omega}(t)= \dot{\theta}(t)$ for every pouring stage determined by \eqref{theta}. We also obtain the lower and upper bounds, ${\omega}_{min}$ and ${\omega}_{max}$, of the angular velocity of the wrist joint's rotation during pouring.  We then use these statistics to form a safety constraint to ensure that the generated angular velocity remains within [${\omega}_{min}$, ${\omega}_{max}$]  during online deployment. 

\subsubsection{Architecture}
The architecture of the fine learning model, denoted $\bm{F_f(.)}$, consists of two  LSTM layers with 128 and 64 units respectively, following the standard described in \cite{hochreiter1997long}. A key aspect of our model is to improve upon previous work using LSTM-based motion control for robotic pouring, by incorporating an attention mechanism \cite{schenck2017visual}. After combining the LSTM model and the attention layer, the generated encoded features pass through a multi-layer perceptron with four hidden units to predict the actions for task execution.

The loss function $\mathcal{L}_{r}$ ($r=\{v_x,v_y,v_z,v_\theta\}$) for each component of the predicted action uses a Mean Square Error (MSE) between the predicted and target outputs. The overall loss function is a sum of the MSE over the four distinct outputs. 

After training the models $F_C(.)$ and $F_f(.)$ using the demonstration database in the source domain $\bm{H}$, the policy $\boldsymbol{\pi}_{{\Theta}}$ can be used for the pouring task in scenarios involving target containers, granules, and background that have been seen before. The domain adaptation in the testing phase is implemented via imaginary learning, which is described next.

\subsection{\textbf{Imaginary Learning}: Domain Adaptation}
\subsubsection{Design Consideration}
The Pix2Pix GAN \cite{isola2017image} has been used for domain adaptation from the simulated environment to a real environment \cite{church2022tactile}. However, aligned image pairs from different domains are required for the model training, which is not realistic for the pouring task because the duration of pouring trajectories in different scenarios may vary significantly. Alterantives such as the
 CycleGAN \cite{zhu2017unpaired}, DiscoGan \cite{kim2017learning} or DualGan \cite{yi2017dualgan} can achieve image translation between different domains using unpaired images in an unsupervised learning manner. Therefore, in this paper, we use CycleGAN to transfer images from the original database to a new database that can then include sufficient synthetic data with the appropriate properties (domain characteristics) for generalization to novel scenarios absent from the original training.
 
Our proposed method will benefit robotic manipulation tasks that are either too difficult to model or too costly to learn from failures via reinforcement learning approaches. 
Unlike prior methods for domain adaptation that require time-aligned and paired demonstrations from different domains to obtain state correspondences, our proposed method enables the robot to adapt to new scenarios via imaginary learning in an unsupervised manner.

 \subsubsection{Implementation and Architecture}
 The imaginary learning phase can be implemented by three steps: i) training the generators via CycleGAN, ii) constructing the synthetic database, and iii) transferring knowledge via fine-tuning.
 
 \textbf{\textit{Step 1:} }\textbf{Training Generators via CycleGAN.}
First, we collect one-shot demonstration data for task execution in a novel scene with new domain characteristic (e.g., a new background, a different container or a different granular material). According to Table \ref{tab:Data1}, scenes S13-S15 have a new background (blue tissue), scene S16-S18 include new granules (brown coffee with irregular shapes), scene S19-S20 include a new container (large black cup), none of which has been demonstrated previously in the training phase. A key aspect of our method is that only one trial of demonstration data need be collected during the testing phase for each of these scenes S13-20, with this database denoted $\mathcal{D}_{\mathcal{T}}^{r}$.

We sample observation data from $\mathcal{D}_{\mathcal{T}}$, which is drawn from the source domain $\bm{H}$, and then sample observation data from $\mathcal{D}_{\mathcal{T}}^{r}$, which is drawn from the target domain $\bm{R}$.
For domain adaptation, we need to train a generator $\bm{G(.)}$ to take observation images from the source domain and  generate new observation images matching those from the target domain, while  a generator $\bm{G'(.)}$ takes the images back from the target domain to the source domain. Meanwhile, a discriminator $\bm{D_H(.)}$ is trained to classify whether a data sample is drawn from the source or from the generated data in the target domain, while a  discriminator $\bm{D_R(.)}$ is trained to classify whether a data sample is drawn from the target or generated data in the source domain. The distribution of the real observations remains fixed, and the distribution of the generating observation is learned to match the real data.  The aim is to solve a min-max problem such that
\begin{equation}
	\label{min-max}
	G^{*}, G^{'*}=\arg \min _{G, G^{'}} \max _{D_{H}, D_{R}} \mathcal{L}\left(G, {G^{'}}, D_{H}, D_{R}\right).
\end{equation}
The details of the loss functions for training the CycleGAN can be found in Appendix.
  
\textbf{\textit{Step 2:} }\textbf{Constructing the Synthetic Database.} The next step is to use the generator $\bm{G(.)}$ to generate new observation data in an imaginary manner to construct a synthetic database $\mathcal{D'}_{\mathcal{T}}^{r}$.
 More specifically, we sample trajectories ${\tau}_1,{\tau}_2,\cdots=[(\bm{o_1},\bm{a_1}),(\bm{o_2},\bm{a_2}),{\cdots}]$, and generate new observation images to replace the original observation images from the original trajectories as new trajectories ${\tau'}_1,{\tau'}_2,\cdots=[(\bm{G}(\bm{o_1}),\bm{a_1}),(\bm{G}(\bm{o_2}),\bm{a_2}), \cdots ]$.  These trajectories are considered to be imaginary demonstration data with generated observation and action pairs, giving $\mathcal{D}_{\mathcal{T}}^{r'}=\{{\tau'}_j\}(j=1,2,\cdots)$ comprised of a series of imaginary trajectories.

\begin{figure}[tb]
	\centering
	\includegraphics[width = 0.9\hsize]{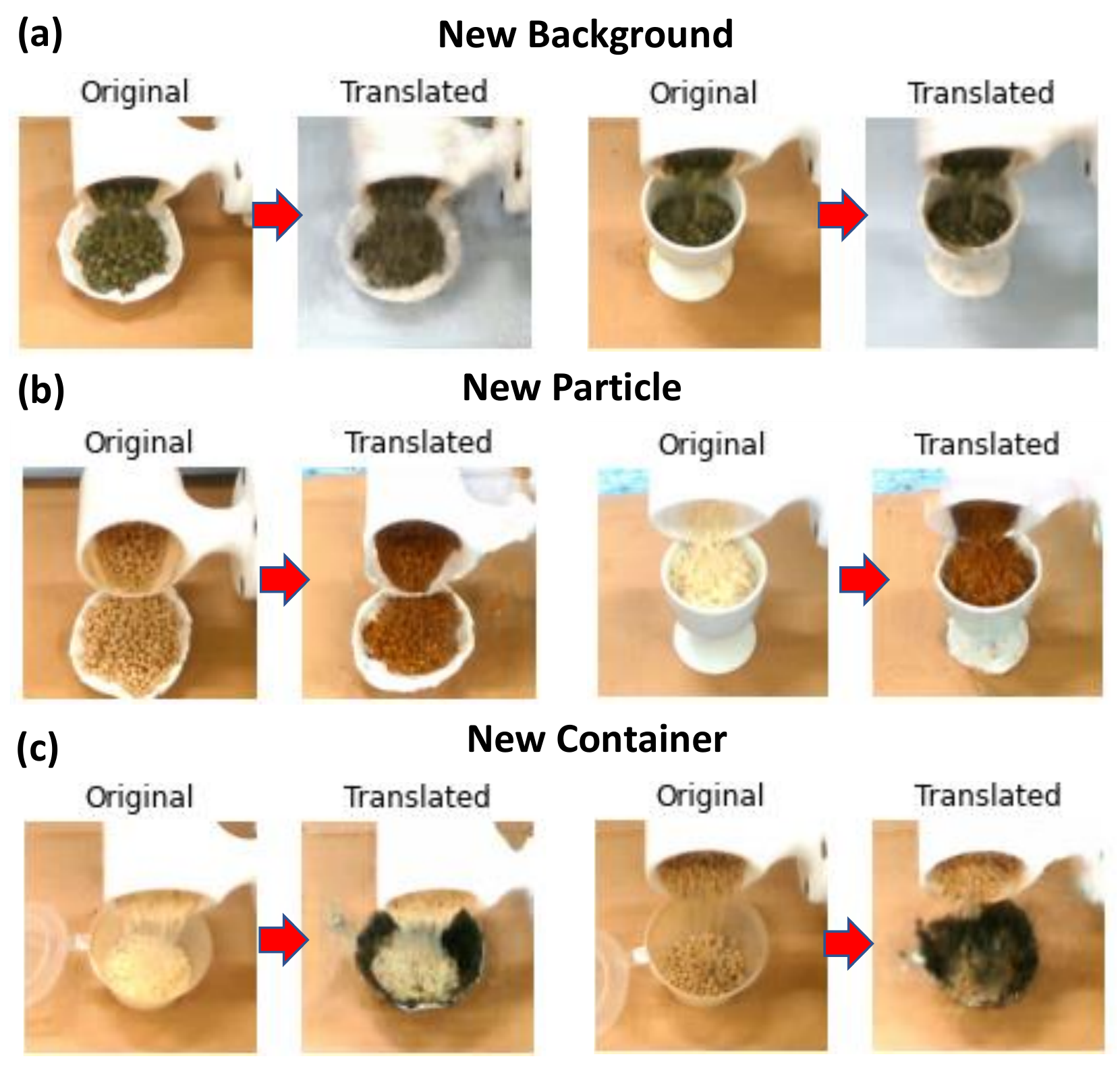}
	\caption{Examples of the results for domain adaptation based on a CycleGan. Image generation to (a) a new background (blue tissue), (b) new granules (coffee with irregular shapes), and (c) a new target container (black cup). }
	\label{fig:Gan}
\end{figure}

 \textbf{\textit{Step 3:} }\textbf{Transferring Knowledge via Fine-Tuning.}
Finally, a new database $\mathcal{D}_{\mathcal{T}}^{o}$ is constructed by combining $\mathcal{D}_{\mathcal{T}}^{r'}$ and $\mathcal{D}_{\mathcal{T}}^{r}$, which is used to fine-tune $\bm{F_C(.)}$ in a transfer learning manner.
Subsequently, we combine $f_p$ (extracted by  $\bm{F_C(.)}$ with updated parameters) and the new task-specific characteristics  $z'$ to generate a new feature vector $V_F^{o}$. In this manner, $\mathcal{D}_{\mathcal{T}}^{o'}$ is constructed by pairing $V_F^{o}$ with the corresponding action values. After fine-tuning $\bm{F_f(.)}$ using $\mathcal{D}_{\mathcal{T}}^{o'}$, a series of generated new policies ${\pi}_{{\Theta}_{J+1}}$, ${\pi}_{{\Theta}_{J+2}}$, $\cdots$ are obtained  for  task execution in novel scenarios with new domain characteristics absent from the original training database $\mathcal{D}_{\mathcal{T}}$ (${z'}\neq{z}$), achieving the appropriate domain adaptation (see Fig. \ref{fig:Gan} for a visualization of image generation using CycleGAN-based techniques to transfer the original image data to synthetic data in the new domain).
 
An overall summary of the workflow for progressive imitation learning is given in \textbf{Algorithm 1}.

\section{Experiments and Results}
\label{experiments}

\subsection{Experiment Design}
 
After initial model training described above, we conducted real-time experiments on a UR5 robot arm, using the setup shown in Fig. \ref{fig:SetupUR1}. 
Three distinct groups of experiments were conducted in total to answer three research questions, which can be known as ablation study for coarse learning, fine learning and imaginary learning respectively.

\begin{itemize}
\item Whether the concept representation features extracted during the coarse learning phase enhance the training efficiency of the action generation model during the fine learning phase or not?

\item How does the performance of the action generation model obtained during the fine learning phase compare with the traditional behavior cloning method in terms of the success rate of the pouring task? 
	
\item Novel domain characteristics represent the usage of new backgrounds, new types of granules, new target containers that have not been included in the database for model training.  Can the algorithm demonstrate its generalizability in the imaginary learning phase by performing the pouring task in new scenarios with  novel domain characteristics? 
\end{itemize}

With the  experiments mentioned above, we can prove the significance of each component for the proposed progressive imitation learning.

\subsection{Coarse Learning Model enhances  data efficiency}
\textbf{This group} of experiments evaluates the effectiveness of the concept representation features for just the coarse learning phase of the overall framework.

An offline analysis is conducted first to evaluate the performance of the coarse learning model.   For an ablation study, we use the one-hot values as features to replace the original concept representation features, so that this comparison study is conducted between with and without  concept representation features.

Without concept representation features, when using the whole database for training the action generation model in the coarse learning phase, the training and testing MSE is 0.0036 and 0.0037 respectively. This is markedly poorer than the training and testing MSE for our proposed method using the extracted features, of 0.0020 and 0.0023 respectively. If 50\% of data from the original database is used for model training, then the  training and testing MSE becomes 0.0022 and 0.0023 respectively, and with only 25\% of data from the original database for model training, the MSE become 0.0024 and 0.0029 respectively. The results are summarized in Table \ref{tab:Results-efficiency}.

\begin{table}[!htb]
	\centering
	\caption{Results for Comparisons between With and Without Coarse Learning and the Analysis of Data Efficiency.}
	\label{tab:Results-efficiency}
	\begin{tabular}{c|cc}
		\hline\hline
		\textbf{Model $\&$ Data}  & \textbf{Training}  & \textbf{Testing}   \\\hline
		\textbf{Without + 100\%Data} & 0.0036 & 0.0037  \\
		\textbf{With + 100\%Data}   &   0.0020 & 0.0023 \\
		\textbf{With + 50\%Data}   &   0.0022 & 0.0023 \\
		\textbf{With + 25\%Data}   &   0.0024 & 0.0029 \\
		\hline\hline
	\end{tabular}
\end{table}

These results show that when using the coarse learning model for concept representation features extraction, the performance of the action generation is better than that without, even if we only use 25\% of the available data. Thus, we conclude that the coarse learning model enhances the data efficiency of the action generation model training.

\subsection{Fine Learning Model improves the success rate}

\textbf{The second group} of
experiments compares the proposed approach with a baseline method which implements imitation learning using one model. The comparison is conducted in terms of the success rate for automatic robotic pouring in four distinct scenes.

\begin{figure}[tb]
	\centering
	\includegraphics[width = 1\hsize]{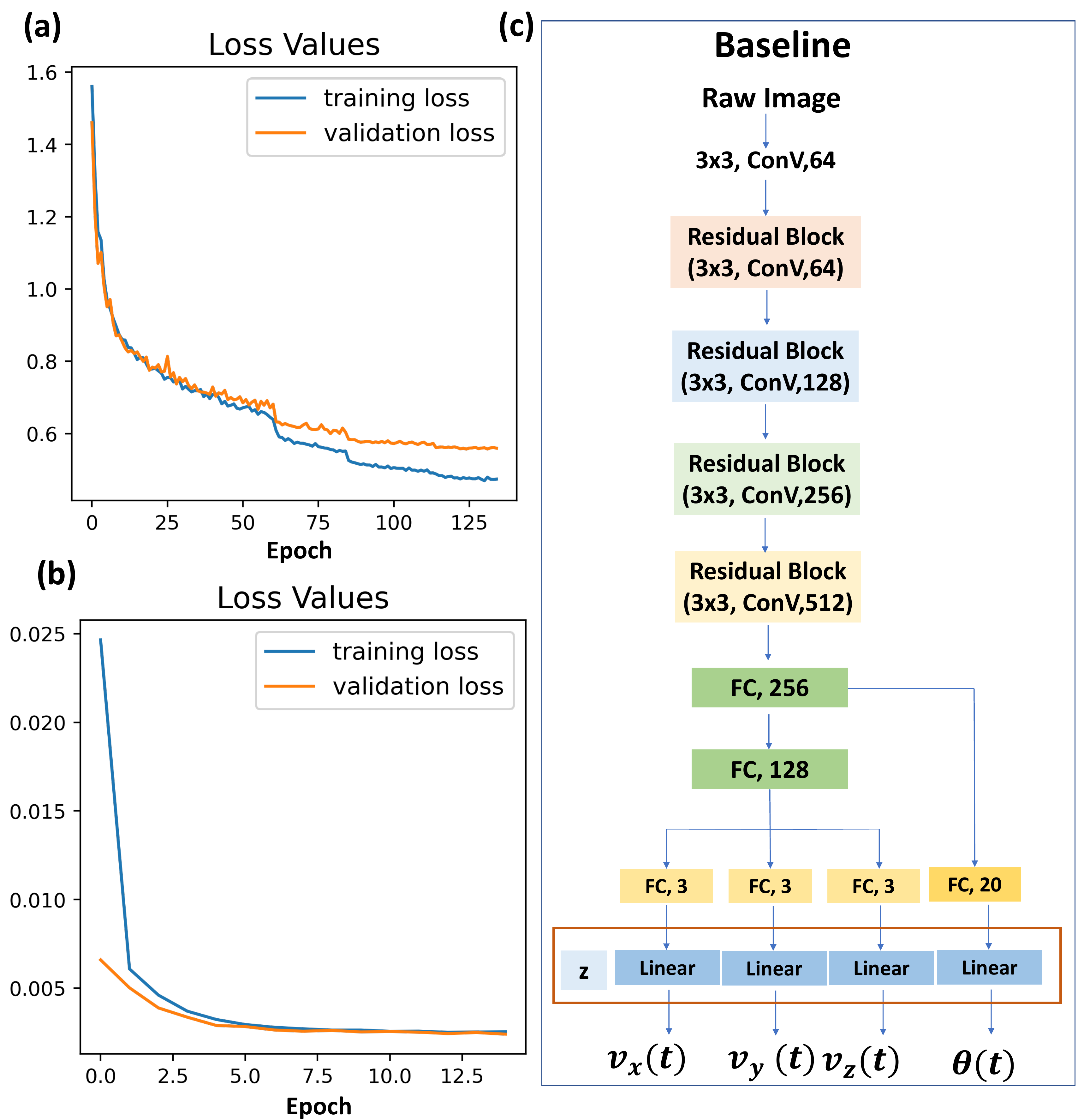}
	\caption{The training and validation loss for (a) the coarse learning model, (b) the fine learning model, (c) baseline model for comparison between with and without fine learning.}
	\label{fig:ExperimentLoss.pdf}
\end{figure}

We used an end-to-end behaviour cloning approach \cite{sharma2018multiple}  for comparison with our proposed progressive learning method to demonstrate the value of fine learning. The neural network architecture for the baseline model is shown in Fig. \ref{fig:ExperimentLoss.pdf}(c).
%\textbf{Appendix D}

\begin{figure}[tb]
	\centering
	\includegraphics[width = 1\hsize]{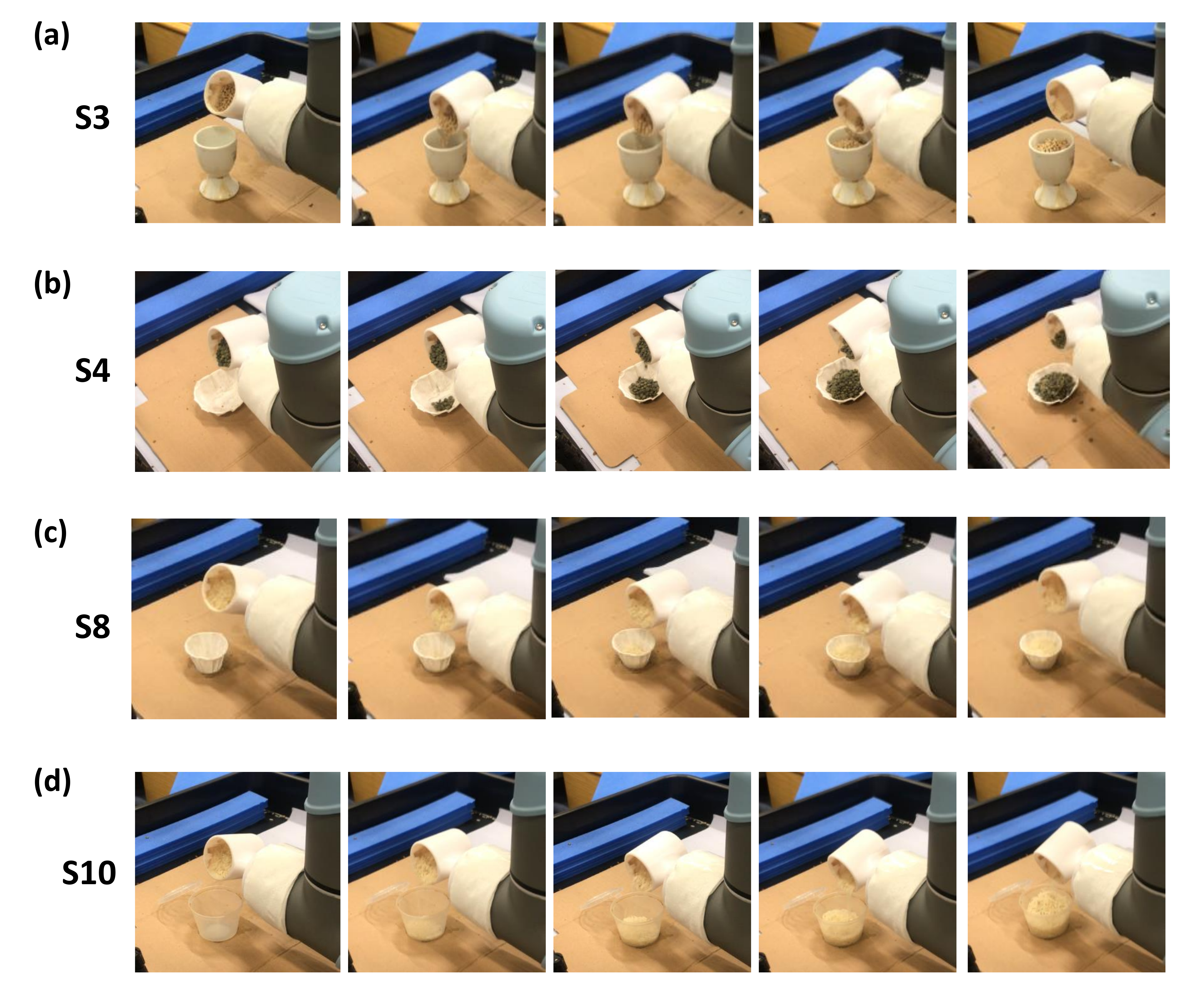}
	\caption{Examples of four successful pouring trials. The experiments were conducted in (a) scene S3; (b)  scene S4; (c)  scene S8; and (d) scene S10 according to Table \ref{tab:Data1}.}
	\label{fig:Experiment-Results.pdf}
\end{figure}
% in the \textbf{Appendix A}##

Scenes S3, S4, S8 and S10 are  selected to cover all the types of granular materials and containers in the experiments. Our evaluation is based on the success rate of the pouring for the different scenes.

\begin{table}[ht]
	\centering
	\caption{Experimental results for comparisons between with and without fine learning. }
	\label{tab:Results}
	\begin{tabular}{c|cccc|c}
		\hline\hline
		& \textbf{S3} & \textbf{S4} & \textbf{S8} & \textbf{S10} &  \textbf{Mean} \\\hline
		\textbf{Without}  &  2/10   & 4/10 &  2/10   &  6/10 & 35.0\%\\
		\textbf{With}  &   6/10   & 8/10 &  7/10   & 10/10 & 77.5\%\\
		\hline\hline		
	\end{tabular}
\end{table}

Fig. \ref{fig:Experiment-Results.pdf} shows four example successful trials of pouring in these scenes, while  Table \ref{tab:Results} summarizes the experimental results on the comparison study. The overall results indicated that with our progressive learning method, the success rate is improved significantly (77.5\% vs. 35.0\%). 
Table \ref{tab:Results}  demonstrates that our proposed method outperforms the traditional end-to-end learning-based model for action generation.

In our view, the reason why the progressive learning method performs better than the baseline method on this task is that the baseline does not utilize temporal information, resulting in the predicted trajectories being unstable and causing the granules to flow out of the container during the pouring process. That said, the success rate for the distinct scenes has differing variances, with the success rate across scenes using the progressing learning method ranging from 60\% to 100\%. This seems due to the fact that some containers have a relatively large opening diameter, and therefore a lower requirement for the precision of pouring motion generation. For those target containers that have smaller opening diameters, the success rate for task execution has the potential for further improvement.

Another inherent advantage of progressive imitation learning is its ability to adapt to new scenes quickly.  The training and validation loss for the coarse learning model $\bm{F_C(.)}$ and the fine learning model $\bm{F_{f}(.)}$ are visualized in Fig. \ref{fig:ExperimentLoss.pdf} (a) and (b) respectively. The training of $\bm{F_C(.)}$ requires more than 100 epochs, while the training of  $\bm{F_{f}(.)}$ only requires 10-20 epochs.  The model training in the fine learning process is much faster than the one in the coarse learning process.  When applied the proposed method to new environment with variances, we can fine-tune the model obtained in fine learning process without retraining the whole model like the end-to-end learning approach.

\subsection{Imaginary Learning ensures the generalizability}

\textbf{This third group} of
experiments is aimed at verifying the generalizability of the progressive learning method. These experiments examine whether the proposed progressive learning method can be adapted to new scenarios with new domain characteristics or not, including new environments (backgrounds), granular materials, and target containers that were not included in the demonstration data collection process for model training.

\begin{figure}[tb]
	\centering
	\includegraphics[width = 1\hsize]{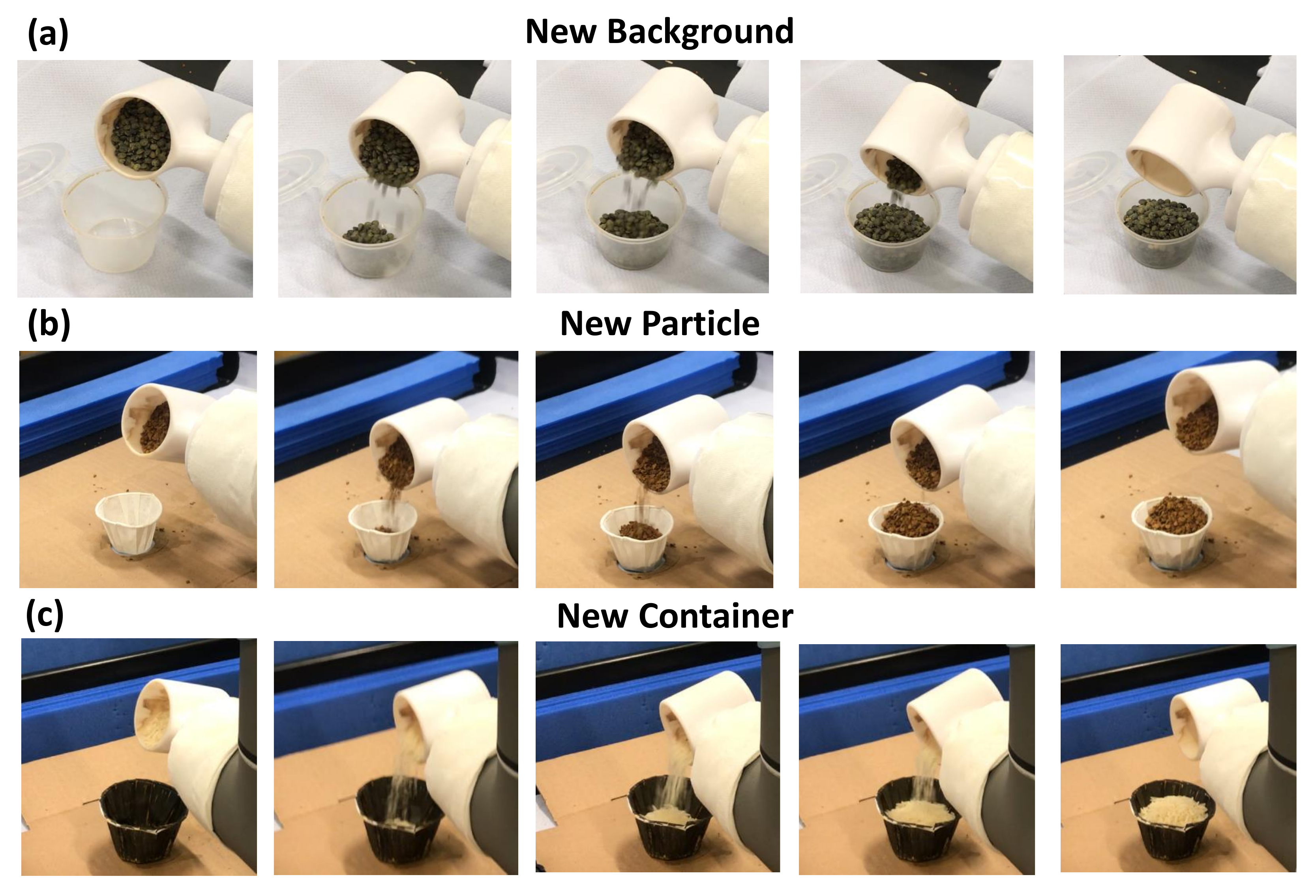}
	\caption{Three successful trials for demonstrating the generalizability of the proposed method. Domain adaptation to (a) a new background, (b) new granules, and (c) a new target container.}
	\label{fig:DifferentCondition.png}
\end{figure}

\begin{table}[!htb]
	\centering
	\caption{Experimental results for comparisons between with and without domain adaptation. }
	\label{tab:Resultsa}
	\begin{tabular}{c|cc}
		\hline\hline
		\textbf{Type of Experiment}  & 	\textbf{Without} & 	\textbf{With}    \\\hline
		\textbf{New Background} &    2/8    & 7/8 \\
		\textbf{New Granules}&     2/8      & 6/8 \\
		\textbf{New Target Container} &    3/8    & 6/8 \\\hline
\textbf{Mean} &    29.1\%    & 79.2\% \\
		\hline\hline
	\end{tabular}
\end{table}
Examples of three successful trials are shown in Fig.~\ref{fig:DifferentCondition.png}, with the experimental results summarized in Table~\ref{tab:Resultsa}.  With the domain adaptation method, the average success rate for the task execution in new scenarios is increased by a large margin from 29.1\% to 79.2\%.  Comparing the success rate with and without domain adaptation, the results indicate that with the adaptation coupled with progressive learning, the robot is able to perform automatic pouring in novel scenarios and has demonstrated good generalizability.

\subsection{Comparisons with Other Work}
The work closest to ours is  in \cite{smith2019avid}, where a  CycleGAN \cite{zhu2017unpaired} is used to translate human demonstrations to robot-looking ones at pixel space. An important difference to this work is that we do not aim to learn a reward function for reinforcement learning and do not require the robot to practice the skill to learn its physical execution. Moreover, in \cite{smith2019avid}, the robot needs to query the human user to indicate success or failure at some critical points during the learning process. We try to avoid  intensive human supervision in this work, and avoid using reinforcement learning that requires trial-and-error for the robot to learn the action generation policy.

To the best of our knowledge, this is the first time that one-shot domain adaptation has been defined.
The one-shot domain-adaptive imitation learning is different from the traditional one-shot learning implemented in  \cite{duan2017one,finn2017one,finn2017model}. For example, the traditional one-shot learning requires the training database to be of high diversity with many demonstration trials for different tasks. For example, about 1300 demonstrations were collected for meta-training during the training phase for \cite{finn2017one}. In our paper, we only need to collect 80 demonstrations, which reduced the need for intensive data collection in the training process.

We compare our proposed progressive learning method with the other methods in terms of performance, data efficiency and generalizability. We assume the algorithm is of high efficiency if the performance does not decrease significantly after reducing 75\% of data for training the model in new scene. If the robot can accomplish a new pouring task with either new environments (backgrounds), granular materials, or target containers that haven't been included in the training dataset, then  the applied method can be known to have good generalizability.  
\begin{table}[!htb]
	\centering
	\caption{Comparisons With Related Work: ``$\checkmark$`` and ``X`` indicate with and without the relevant properties, respectively.}
	\label{tab:Results-compare}
	\begin{tabular}{c|c|c|c}
		\hline
	\multirow{ 2}{*}{\textbf{Methods}} 	& \textbf{Performance}&\multirow{ 2}{*}{\textbf{Data Efficiency}} & \multirow{ 2}{*}{\textbf{Generalizability}} \\
		& (MSE)  && \\\hline
		%\textbf{Hard-Code} & /  & 0.05   & /& X \\		
		\textbf{BC-LSTM}\cite{chen2019accurate} &  0.0023  & $\checkmark$ & X\\
		\textbf{BC-DCNN}\cite{zhang2018deep} & 0.0158 & X & X\\
		\textbf{BC-ResNet}\cite{zhang2021explainable} & 0.0037  & X & X  \\
	  % \textbf{TCN} &  Medium & Medium & 0.35 &  $\checkmark$ & X & X & \cite{sermanet2017time} \\
	%	\textbf{MAML}&  Large & Small & / & / & $\checkmark$ & $\checkmark$ &   \cite{finn2017model}  \\
		\textbf{Proposed} &  0.0023 & $\checkmark$ & $\checkmark$  \\
		\hline
	\multicolumn{4}{c}{	``/`` indicates not applicable.}
	\end{tabular}
\end{table}

Table~\ref{tab:Results-compare} shows the comparisons of the proposed method with other related work that can be applied to the automatic pouring task, including deep imitation learning method (also known as behavior cloning) based on LSTM, deep convolutional neural network (DCNN)~\cite{zhang2018deep}, deep residual neural network (ResNet) respectively \cite{zhang2021explainable}.  We can conclude that our proposed method has the advantage of high success rate, data efficiency and generalizability.

\subsection{Discussion}

\subsubsection{Adaptation to Other Tasks}

The purpose of this work is to provide a general framework that can bring benefits to automation in both service and industrial robotics.  While our experiments focus on a robotic pouring task, our framework is not specific to this task, and could also be used for a wide variety of similarly dexterous tasks, spanning from making consumer beverages to handling dangerous chemical solvents, building a pyramid of cups, washing dishes, opening doors.
Take door opening task as an example, in the coarse learning phase, we can replace the component of `tilt angle control' with the `door opening angle control', while the component of 3D position adjustment remains the same. As for the task characteristics, we can replace the various types of granules with the various types of door handles' shapes. The fine learning and imaginary learning phase can remain the same.

%\textcolor{red}{extend to 6DoF control}
%\textcolor{red}{concrete example of tactile-based pushing}
%Though this paper focuses on the robotic pouring task, the proposed method can be extended to a series of sequential tasks, such as opening doors, washing dishes, , etc.

Moreover, the proposed framework could be modified to more complex scenarios that require the robot to learn from demonstration based on sensory input with multiple modalities. For example,  we use image data as observations in this work, while force/torque sensors, tactile sensors, RGB-D/stereo cameras could be incorporated into the proposed progressive learning approach to provide more comprehensive and accurate information as observations, further enhancing the efficiency and generalizability of robot learning. 

\subsubsection{Future Work}
To probe limitations of the approach, we notice that for the adaptation to new backgrounds, the lighting condition may influence slightly the performance of the model. In particular, for a background material with high reflectivity, the success rate of the pouring task was reduced, which we attribute to the strong reflected light leading to poor performance of the robot's perception system. Moreover, were the granules to have the same color as the target container, then the pouring task may become more challenging. In the future, we will enhance the robustness of the proposed method by incorporating more sensory information with effective feature extraction techniques.

%combine force, tactile and vision information as a multi-modality sensing system to provide observation data for the robot to enhance the robustness of the proposed method by incorporating more sensory information.

The proposed progressive learning framework include three essential modules, each of which can be further improved. For example, we can use more efficient representation learning approach to generate the concept representation for downstream tasks, which can further accelerate the training speed of the fine learning model and enhance its performance. As for  fine learning, we can employ Neural Architecture Search (NAS) to optimize the neural network model, which has high potential to outperform manually designed models. 

Imaginary learning is the key step towards  domain adaptation to new scenarios. CycleGAN is deployed 
to generate new data in this paper, which can be regarded as a plug-in function. The application of CycleGAN in our proposed progressive imitation learning method is novel, since it enables imaginary learning for adaptation to unseen scenarios. However, the current limitation is that we need to retrain the generator when new scenario shows up. To this end, other types of GANs such as  Lifelong GAN \cite{zhai2019lifelong} can be used in the future development to update the generator  continuously for adaptation to new task. 
%transfer learned knowledge from previous trained generators to a new generator

%\textcolor{blue}{For section 3, how much did the success rate for the new granules depend on the similarity of original dataset that was used as a starting point? Did the original dataset need to contain a wide variety of different object materials that may result in different observation/action pairs to capture the effect of different types of materials?“-----A wide variety of different materials in the original database may benefit the enhancement of the success rate after domain adaptation. However, we have proved in the manuscript that the algorithm can ensure a high success rate when we pour unseen granules into unseen target containers. We will add clarification that the proposed method does not require high similarity between the materials used in the training phase and the testing phase.}

\section{Conclusions}
In this paper, we proposed a  progressive learning framework to achieve one-shot domain adaptive imitation learning.  The robotic manipulator learned a concept representation in the coarse learning phase, and then progressed to learning how to generate accurate actions for task execution across multiple pouring scenarios in the fine learning phase. To improve the generalization to new domains, imaginary learning is implemented via  CycleGAN to generate new observations for the robot to enhance its perception capability in new scenarios during the imaginary learning phase. 

%In this way, the robot can learn to have a high success rate for task execution in new scenarios after observing a one-shot human demonstration. 

Our experiments were based on a robotic pouring task that required the robot to pour different types of granular materials into distinct target containers in front of different backgrounds. We  verified that our proposed method has advantages in terms of high \textbf{success rate}, \textbf{data efficiency} and \textbf{generalizability}. 
With our progressive learning method,  the success rate for the robotic pouring task can reach 77.5\%, while the performance of the fine learning model can maintain even if we only use 25\% of data for training. Moreover, with imaginary learning phase, the robot can increase the success rate for task execution in new scenarios from 29.1\% to 79.2\%.

To the best of our knowledge, this is the first time that one-shot domain adaptation has been carried out for a robotic pouring task in a physical environment. Moreover, we proposed the progressive learning method, which can be widely applied to different learning-based robotic manipulation tasks.

%More advanced models can be developed in the future when considering more complex tasks.

\section*{Appendix}

%\subsection{Details for the Training Database}

Suppose that $n$ is the total number of samples used for calculating the loss function, the adversarial loss on the observation samples in domain $\bm{R}$ can be calculated as follows:
\begin{equation}
	\begin{aligned}
		\label{dis}
		\mathcal{L}_{\mathrm{adv}}\left(\bm{G}, \bm{D_{R}}, \bm{H}, \bm{R}\right)=\frac{1}{n} \sum_{i=1}^{n}\left(\bm{D_{R}}\left(x_i^r\right)-1\right)^{2}+\\
		\frac{1}{n} \sum_{i=1}^{n}\left(\bm{D_{R}}\left(\bm{G}\left(x_i^h\right)\right)\right)^{2}
	\end{aligned}
\end{equation}

Similarly, the adversarial loss on the observation samples in domain $\bm{H}$ can be calculated as follows:
\begin{equation}
	\label{dis1}
	\begin{aligned}
		\mathcal{L}_{\mathrm{adv}}\left({\bm{G^{'}}}, \bm{D_{H}}, \bm{R}, \bm{H}\right) =
		\frac{1}{n} \sum_{i=1}^{n}\left(\bm{D_{H}}\left(\bm{G^{'}}\left(x_i^r\right)\right)\right)^{2}+\\
		\frac{1}{n}\sum_{i=1}^{n}\left(\bm{D_{H}}\left(x_i^h\right)-1\right)^{2}
	\end{aligned}
\end{equation}

The cycle consistency loss can be calculated as follows.

\begin{equation}
	\label{dis2}
	\begin{aligned}
		\mathcal{L}_{\text {cyc }}(\bm{G}, \bm{G^{'}})=\mathcal{L}_{\text {cyc}}^{1} + \mathcal{L}_{\text {cyc }}^{2}  &=\frac{1}{n}\sum_{i=1}^{n}\left[\|\bm{G^{'}}(\bm{G}(x_i^h))-x_i^h\|_{1}\right] \\
		&+\frac{1}{n}\sum_{i=1}^{n}\left[\|\bm{G}(\bm{G}^{'}(x_i^r))-x_i^r\|_{1}\right]
\end{aligned}\end{equation}
where $\vert\vert{.}\vert\vert$ represents the L1 norm (Manhattan norm). The overall loss is computed by adding the adversarial loss of  $\bm{G(.)}$ and $\bm{G^{'}(.)}$ as well as the cycle consistency loss, which is defined as follows:

\begin{equation}
	\label{dis3}
	\begin{aligned}
		\mathcal{L}\left(\bm{G}, \bm{G^{'}}, \bm{D_{H}}, \bm{D_{R}}\right) &=\mathcal{L}_{\mathrm{adv}}\left(\bm{G}, \bm{D_{R}}, \bm{H}, \bm{R}\right) \\
		&+\mathcal{L}_{\mathrm{adv}}\left(\bm{G^{'}}, \bm{D_{H}}, \bm{R}, \bm{H}\right) \\
		&+\lambda \mathcal{L}_{\mathrm{cyc}}(\bm{G}, \bm{G^{'}})
\end{aligned}\end{equation}
where $\lambda$ is a parameter that controls the relative importance between the adversarial loss and the cycle consistency loss. The target is to solve a min-max problem as follows:
\begin{equation}
	\label{min-max1}
	G^{*}, G^{'*}=\arg \min _{G, G^{'}} \max _{D_{H}, D_{R}} \mathcal{L}\left(G, {G^{'}}, D_{H}, D_{R}\right)
\end{equation}

\bibliographystyle{IEEEtran}
\bibliography{references}

\clearpage
\newpage

\end{document}